# Reconstructing the somatotopic organization of the corticospinal tract remains a challenge for modern tractography methods

Jianzhong He[1], Fan Zhang[2]*, Yiang Pan[1], Yuanjing Feng[1]*, Jarrett Rushmore[5,6], Erickson Torio[4], Yogesh Rathi[2,3], Nikos Makris[3,5], Ron Kikinis[2], Alexandra J. Golby[2,4], Lauren J. O'Donnell[2]

[1] Institution of Information Processing and Automation, Zhejiang University of Technology, Hangzhou, China
[2] Department of Radiology, Brigham and Women's Hospital, Harvard Medical School, Boston, USA
[3] Department of Psychiatry, Brigham and Women's Hospital, Harvard Medical School, Boston, USA
[4] Department of Neurosurgery, Brigham and Women's Hospital, Harvard Medical School, Boston, USA
[5] Departments of Psychiatry, Neurology and Radiology, Massachusetts General Hospital, Harvard Medical School, Boston, USA
[6] Department of Anatomy and Neurobiology, Boston University School of Medicine, Boston, Massachusetts, USA

* Co-corresponding-authors

## Abstract

The corticospinal tract (CST) is a critically important white matter fiber tract in the human brain that enables control of voluntary movements of the body. The CST exhibits a somatotopic organization, which means that the motor neurons that control specific body parts are arranged in order within the CST. Diffusion MRI tractography is increasingly used to study the anatomy of the CST. However, despite many advances in tractography algorithms over the past decade, modern, state-of-the-art methods still face challenges. In this study, we compare the performance of six widely used tractography methods for reconstructing the CST and its somatotopic organization. These methods include constrained spherical deconvolution (CSD) based probabilistic (iFOD1) and deterministic (SD-Stream) methods, unscented Kalman filter (UKF) tractography methods including multi-fiber (UKF2T) and single-fiber (UKF1T) models, the generalized q-sampling imaging (GQI) based deterministic tractography method, and the TractSeg method. We investigate CST somatotopy by dividing the CST into four subdivisions per hemisphere that originate in the leg, trunk, hand, and face areas of the primary motor cortex. A quantitative and visual comparison is performed using diffusion MRI data (N=100 subjects) from the Human Connectome Project. Quantitative evaluations include the reconstruction rate of the eight anatomical subdivisions, the percentage of streamlines in each subdivision, and the coverage of the white matter-gray matter (WM-GM) interface. CST somatotopy is further evaluated by comparing the percentage of streamlines in each subdivision to the cortical volumes for the leg, trunk, hand, and face areas. Overall, UKF2T has the highest reconstruction rate and cortical coverage. It is the only method with a significant positive correlation between the percentage of streamlines in each subdivision and the volume of the corresponding motor cortex. However, our experimental results show that all compared tractography methods are biased toward generating many trunk streamlines (ranging from 35.10%-71.66% of total streamlines across methods). Furthermore, the coverage of the WM-GM interface in the largest motor area (face) is generally low (under 40%) for all compared tractography methods. Finally, other than UKF2T, all tractography methods produce a negative correlation between the percentage of streamlines in each subdivision and the volume of the corresponding motor cortex, indicating that reconstruction of CST somatotopy is still a large challenge. Overall, we conclude that while current tractography methods have made progress toward the well-known challenge of improving the reconstruction of the lateral projections of the CST, the overall problem of performing a comprehensive CST reconstruction, including clinically important projections in the lateral (hand and face area) and medial portions (leg area), remains an important challenge for diffusion MRI tractography.

## 1. Introduction

The pyramidal or corticospinal tract (CST) is a large descending white matter motor pathway that carries motor signals from the cortex and is critical for human voluntary movement. The corticobulbar tract enables face, head, and neck movements such as speech and swallowing, and it is often studied as part of the corticospinal tract. (In this study, tractography seeding includes the corticospinal and corticobulbar tracts.) The CST originates in several regions, including the primary motor cortex (about 60% of CST fibers (Banker & Tadi, n.d.; Dalamagkas et al., 2020; Davidoff, 1990; Jane et al., 1967)), the supplementary motor cortex, and the parietal lobe (Davidoff, 1990; Levin & Beadford, 1938; Nyberg-Hansen & Rinvik, 1963; Russell & DeMyer, 1961). The CST fibers originating in the primary motor cortex have a somatotopic organization related to the motor homunculus (Ghimire et al., 2021; Gordon et al., 2023; Roux et al., 2020), the topographic representation of the human body in the motor cortex. In this paper we focus on four main areas of the motor cortex responsible for movements of the leg (lower extremity), trunk, hand (upper extremity), and face (head and neck). See Figure 1 for an anatomical overview of the CST.

The CST is often studied using diffusion MRI tractography, the only method that allows in-vivo, non-invasive mapping of the human brain's white matter connections (Basser et al., 2000). CST tractography is clinically used in neurosurgical planning to avoid injury to the corticospinal tract (Berman et al., 2004; Bucci et al., 2013; Essayed et al., 2017a; Farquharson, Tournier, Calamante, et al., 2013) and to predict motor outcomes after stroke (Nguyen et al., 2022). Furthermore, CST tractography enables the study of the anatomy and variability of this critical motor pathway in human health (Dalamagkas et al., 2020; Imfeld et al., 2009) and disease (Ellis et al., 1999; Groisser et al., 2014; Jung et al., 2022; Kirton et al., 2007; Kwon et al., 2016). The somatotopic organization of CST tractography has been studied at the level of the brainstem (J. H. Hong et al., 2010; Park et al., 2008), the posterior limb of the internal capsule (Holodny, Gor, et al., 2005; Holodny, Watts, et al., 2005; Ino et al., 2007; Kim et al., 2008; Pan et al., 2012; Park et al., 2008), and the corona radiata (Jang & Seo, 2015; Kwon et al., 2014), while functional magnetic resonance imaging and transcranial magnetic stimulation have been used to define motor cortical regions of interest for tractography (Bucci et al., 2013; Conti et al., 2014; Qazi et al., 2009; Weiss et al., 2015). Over the past two decades, many diffusion MRI (dMRI) tractography methods have been proposed for the reconstruction of the CST, including single- and multi-fiber modeling, deterministic and probabilistic methods, and machine-learning-based methods.

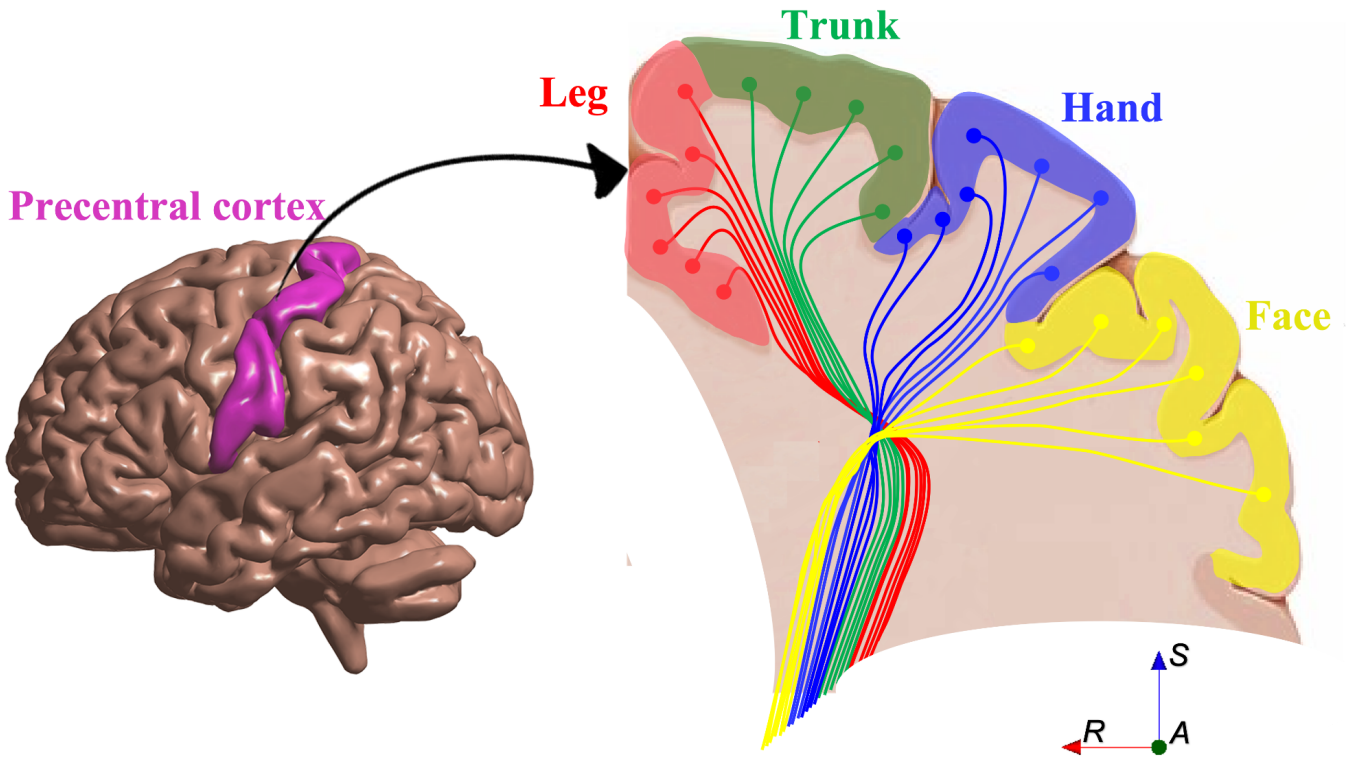

Figure 1. A schematic anatomical overview of the CST and its somatotopic organization (Ghimire et al., 2021; Pan et al., 2012). There are four major subdivisions shown, face fibers (yellow), hand fibers (blue), trunk fibers (green), and leg fibers (red). These subdivisions are responsible for controlling corresponding body muscles.

Despite many recent advances, dMRI tractography of the CST still faces challenges in adequately depicting all portions of the CST (Petersen & McIntyre, 2023; Pujol et al., 2015; Zhylka et al., 2023). An ideal tractography reconstruction of the CST should produce streamlines that cover (intersect) all regions of the primary motor cortex. Furthermore, based on anatomical studies in non-human primates, the neuron density is higher in the lateral than the medial primary motor cortex (Nudo & Masterton, 1990; Young et al., 2013), a finding corroborated by recent stereological analysis in the human primary motor cortex (Rivara et al., 2003). Therefore, an ideal tractography reconstruction of the CST should reconstruct a higher percentage of streamlines from lateral motor areas in comparison to medial motor areas. In comparison with single-fiber tractography, multi-fiber tractography can improve reconstruction of the lateral part of the CST that originates in face motor areas (Akter et al., 2011; Pan et al., 2012; Raffa et al., 2018; Yamada et al., 2007). However, adequate reconstruction of the CST, especially the lateral motor fibers, is still impeded by dMRI tractography challenges including gyral bias (Rheault et al., 2020; K. Schilling et al., 2018; Wu et al., 2020), crossing fibers (Alexander & Seunarine, n.d.; Maier-Hein et al., 2017; Tournier, n.d.; Tuch et al., 2003), and bottleneck regions (Girard et al., 2020; Maier-Hein et al., 2017; K. G. Schilling et al., 2022). For example, Figure 2 gives a visualization of a crossing fiber region where the corpus callosum, SLF II-IV and CST cross in the centrum semiovale, and a bottleneck fiber region where the subdivisions of leg, trunk, hand, and face converge throughout a local region and share the same orientation. In addition, another known challenge for CST reconstruction is that different tractography methods may give different results (Rheault et al., 2020; K. G. Schilling et al., 2021).

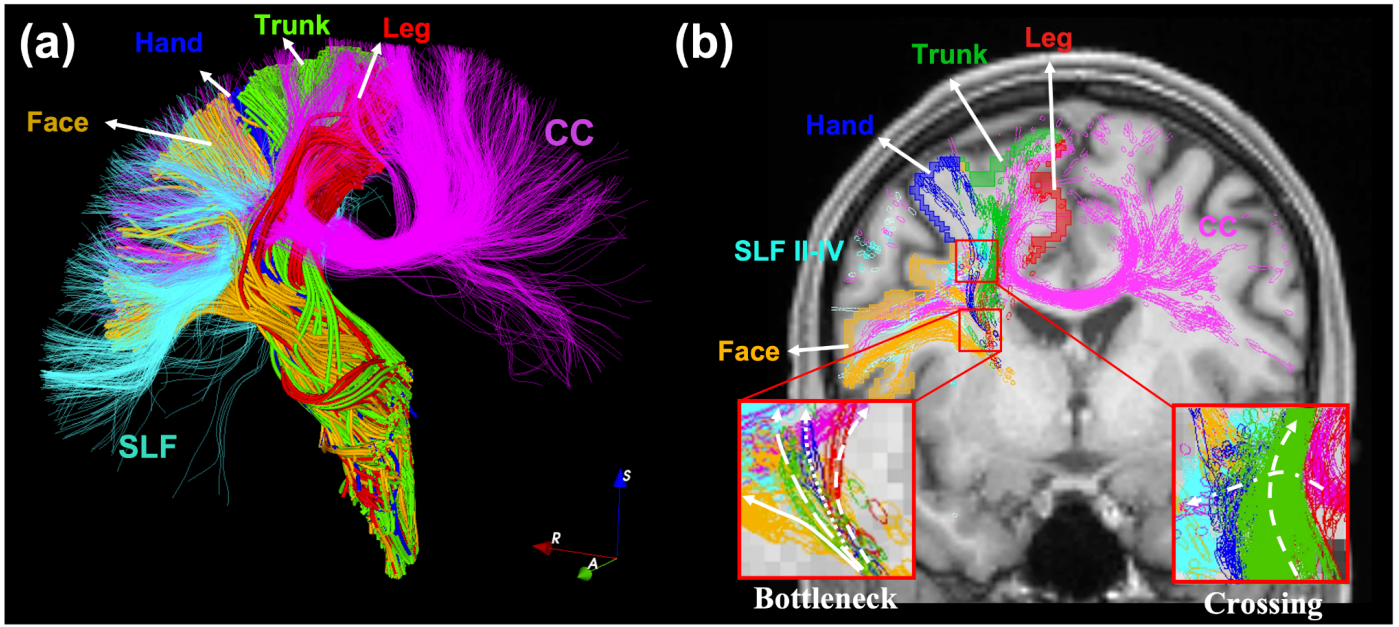

**Figure 2.** Crossing region of the Corpus Callosum (pink), the Superior Longitudinal Fasciculus (cyan) and the CST in the centrum semiovale. (a) 3D oblique superior-anterior view of bundle streamlines and four motor areas, (b) Oblique coronal cross section of five bundle streamlines.

The main contribution of this work is to comprehensively investigate the performance of several tractography methods for reconstructing the CST and its somatotopic organization at the level of the motor cortex. In particular, we investigate anatomical subdivisions of the CST by dividing it into four parts in each hemisphere, originating in leg, trunk, hand, and face areas of the primary motor cortex. In this work, six different tractography methods are compared, including two methods based on the constrained spherical deconvolution (CSD) model (J. D. Tournier et al., 2010; J.-D. Tournier et al., 2012), two methods that use the unscented Kalman filter (UKF) tractography framework (Liao et al., 2017; Malcolm et al., 2010; Reddy & Rathi, 2016), one method based on the Generalized Q-Sampling imaging (GQI) model (F. Yeh et al., 2010; F.-C. Yeh et al., 2013), and a deep learning based tractography method (Wasserthal et al., 2019). A comprehensive quantitative and visual comparison is performed using dMRI data (N=100 subjects) from the Human Connectome Project (HCP) Young Adult data repository (Glasser et al., 2013; Van Essen et al., 2013). Quantitative evaluation includes the reconstruction rate of the eight anatomical areas, the percentage of streamlines in each anatomical subdivision, and the coverage of white matter-gray matter (WM-GM) interface. The quality of the somatotopic reconstruction is then evaluated by comparing the anatomical distribution of reconstructed streamlines to cortical volume measures for leg, trunk, hand, and face areas. Furthermore, to assess the reliability of the CST tracking results, a population-based CST visualization is performed to compare the spatial overlap of CSTs reconstructed.

## 2. Materials and Methods

### 2.1 Evaluation datasets

We used MRI data from a total of 100 subjects (the "100 Unrelated Subjects" release) (54 females and 46 males, age: 22 to 35 years old) from the Human Connectome Project (HCP) database (https://www.humanconnectome.org) (Van Essen et al., 2013) for experimental evaluation. The HCP scanning protocol was approved by the local Institutional Review Board (IRB) at Washington University. The HCP database provides dMRI data acquired with a high-quality image acquisition protocol using a customized Connectome Siemens Skyra scanner (Sotiropoulos et al., 2013; Uğurbil et al., 2013) and processed using a well-established processing pipeline (Glasser et al., 2013) including motion correction, eddy current correction, EPI distortion correction, and co-registration between dMRI and T1-weighted (T1w) data. The dMRI acquisition parameters in HCP were: TR=5520 ms, TE=89.5 ms, FA=78°, voxel size=1.25×1.25×1.25 mm$^3$, and FOV=210×180 mm$^2$. A total of 288 images were acquired in each dMRI dataset, including 18 baseline images with a low diffusion weighting of b=5 s/mm$^2$ and 270 diffusion-weighted images evenly distributed at three shells of b=1000/2000/3000 s/mm$^2$. In our study, we used the single-shell 3000 s/mm$^2$ data to perform CST fiber tracking for different tractography methods as applied in our previous studies (He et al., 2021; Zhang, Wu, et al., 2018). We choose that b3000 data because angular resolution is better and more accurate at high b-values (Descoteaux et al., 2007; Ning et al., 2015). In addition, we also used the anatomical T1w data (co-registered with the dMRI) to facilitate creation of tractography seeding masks and cortical and subcortical regions of interest (ROIs) for streamline selection. The T1w image acquisition parameters were: TR=2400 ms, TE=2.14 ms, and voxel size=0.7×0.7×0.7 mm$^3$. More detailed information about the HCP data acquisition and preprocessing can be found in (Glasser et al., 2013).

### 2.2 Comparison of six tractography methods

To investigate which tractography toolkits have been used in CST reconstruction in the literature, a PubMed search was performed using the terms "pyramidal tract" or "corticospinal tract" or "CST," and "tractography" or "fiber tracking." Then, we focused on studies that reported tractography of the CST in humans, provided detailed tracking parameters, described a streamline selection strategy, and displayed CST bundle rendering. From these studies, we selected three representative tractography toolkits that implement methods that have been widely used in CST reconstruction. These are MRtrix3[1] (J.-D. Tournier et al., 2012, 2019), DSI Studio[2] (F. Yeh et al., 2010; F.-C. Yeh et al., 2013), and SlicerDMRI[3] (Norton et al., 2017; Zhang, Noh, et al., 2020). The tractography methods (shown in Table 1) from these toolkits have shown good performance in CST reconstruction and are computationally efficient. Furthermore, we included a popular automated tractography toolkit called TractSeg[4] (Wasserthal et al., 2018, 2019), which has been shown to be highly effective in reconstructing the CST using deep learning. We note that our literature search demonstrated the high popularity of DTI-based tractography using a single tensor model; however, due to the well-known limitations of DTI in tracing the CST (Maier-Hein et al., 2017; K. G. Schilling, Nath, et al., 2019), in this study we only include one DTI-based tractography method (1-tensor UKF) for comparison.

---

[1] https://www.mrtrix.org/
[2] http://dsi-studio.labsolver.org/
[3] http://dmri.slicer.org/
[4] https://github.com/MIC-DKFZ/TractSeg

| Table 1. Tractography methods studied in this work. | | |
|---|---|---|
| **Toolkit** | **Methods (model + tractography)** | **CST research using toolkit** |
| MRtrix3 | MSMT-CSD + iFOD2 | (Bech et al., 2018; Brusini et al., 2019; Daducci et al., 2016; Farquharson, Tournier, & Calamante, 2013; Konopleva et al., 2019; Reid et al., 2016, 2017; K. G. Schilling, Gao, et al., 2019; Stefanou et al., 2016; Wu, Feng, et al., 2018; Zhylka et al., 2020) |
| | MSMT-CSD + SD-Stream | (Aydogan & Shi, 2018; Barakovic et al., 2021; Cousineau et al., 2017; Garyfallidis et al., 2015, 2018; He et al., 2019) |
| DSI Studio | GQI + Deterministic (GQI-Det) | (Jiang et al., 2020; F.-C. Yeh et al., 2018) |
| SlicerDMRI | 1-tensor + UKF (UKF1T) | (Z. Chen et al., 2015, 2016; Liao et al., 2017; Zhang, Noh, et al., 2020) |
| | 2-tensor + UKF (UKF2T) | (Z. Chen et al., 2015, 2016; Dalamagkas et al., 2020; Essayed et al., 2017b; Feng et al., 2020; Sydnor et al., 2018; Wu, Zhang, et al., 2018; Zhang, Cetin Karayumak, et al., 2020; Zhang, Savadjiev, et al., 2018; Zhang, Wu, et al., 2018) |
| TractSeg | Tract orientation mapping + bundle-specific tractography (TractSeg) | (Bryant et al., 2021; Fuelscher et al., 2021; Wasserthal et al., 2020) |

### 2.3 Regions of interest for tractography and selection of CST

CST tractography was conducted within a mask, which was larger than the possible region through which the CST passes. This procedure enabled comprehensive CST reconstruction while restricting to the potential CST region for efficiency, as applied in our previous tractography studies (Xie et al., 2020; Zhang, Xie, et al., 2020). Each mask was automatically created based on a FreeSurfer segmentation (Fischl, 2012). The mask image started from the anterior slice of the precentral cortex to the posterior slice of the brainstem at the coronal view (shown in Figure 3) to fully cover the potential CST pathway. We visually checked the computed mask for each subject.

For seeding of tractography, appropriate seed regions within the mask (whole brain, WM, GM or WM-GM interface) were employed for each tractography method (details about the selected region for each method are described in Section 2.4 under *Considerations for fair comparison across methods*). After seeding tractography, an exclusion ROI corresponding to the midsagittal plane was used in order to guarantee that the CST streamline remained ipsilateral. To do so, a manually drawn midsagittal ROI in the MNI space was transformed to each subject's DWI data space, via an image registration between the MNI T1w data and the subject-specific b0 image, as shown in Figure 4(a). For each subject, streamlines passing through the midsagittal plane were removed. Finally, a streamline length threshold of 70mm was applied to eliminate any effect from streamlines too short to form part of the CST anatomy (Faraji et al., 2015; Javadi et al., 2017).

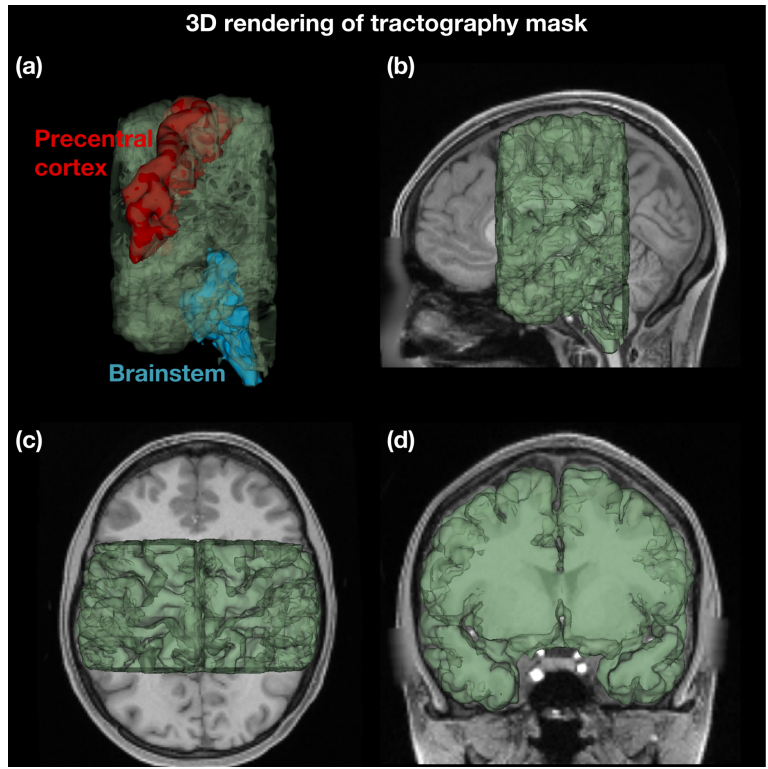

Figure 3. Each mask for tractography (green surface) fully covers the precentral cortex (red surface) and brainstem (blue surface). (b), (c), and (d) are the 3D renderings of the mask from the sagittal, axial, and coronal view, respectively.

To select and identify the CST, we used ROI-based streamline selection (Bech et al., 2018; Dalamagkas et al., 2020; Ille et al., 2021; Nemanich et al., 2019). We extracted all streamlines connecting the precentral cortex and the brainstem (ROIs shown in Figure 4(b)). We identified subdivisions of the CST using our modification of the Desikan-Killiany cortical atlas (Desikan et al., 2006) that includes expert-defined delineations of the hand, face, trunk, and leg motor areas (Y. Hong et al., 2018). FreeSurfer software[5] was used to parcellate cortical motor areas using the expert-defined atlas. Finally, the CST streamlines were separated into eight subdivisions by selecting streamlines that connected the eight motor areas (hand, face, trunk, and leg motor areas in the left and right hemispheres) and the brainstem for each individual subject.

---

[5] http://surfer.nmr.mgh.harvard.edu/

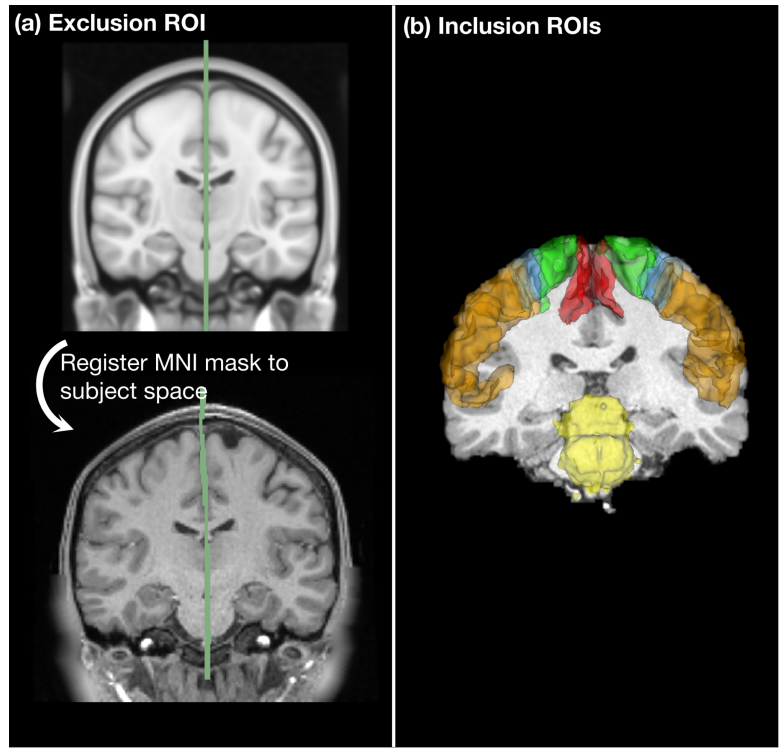

Figure 4. Exclusion ROI. Streamlines passing through the middle green line will be removed. (b) Inclusion ROIs: The brainstem region (yellow) and the precentral cortex, which includes four anatomical motor areas from each hemisphere (red: leg, green: trunk, blue: hand, orange: face).

### 2.4 Experimental evaluation

We performed CST tractography and subdivision identification in the 100 subject dataset as described above. The performance of all methods was quantified with the following metrics.

**Metric 1: Reconstruction rate of complete CST**

This metric assesses the ability of each tractography method to successfully reconstruct all eight subdivisions (4 subdivisions per hemisphere) of the CST. The successful streamline reconstruction or identification rate has previously been applied in several studies to quantify tractography performance (Côté et al., 2013; He et al., 2021; Maier-Hein et al., 2017). Any subdivision with fewer than 10 streamlines was marked as not successfully traced after ROI selection. This fiber threshold has been applied in many studies (Geisler et al., 2022; Rathi et al., 2011). The reconstruction rate of the complete CST pathway was computed as the percentage of subjects where all eight anatomical subdivisions (2 hemispheres × 4 anatomical motor areas) were successfully reconstructed. **Statistical analysis:** A statistical comparison was then performed across the six tractography methods. The reconstruction rate was encoded using the number 1 if all subdivisions were detected and 0 indicated if at least one subdivision was not found. Cochran's Q test (Cochran, 1950) was then performed across the six methods. Following that, McNemar's tests (McNEMAR, 1947) were performed between each pair of the six tractography methods (a total of 15 comparisons) as a post-hoc test. These statistics are appropriate for nominal data and have been applied in many studies (Leavens et al., 2004; Ludwig & Simner, 2013; Tietäväinen et al., 2015).

**Metric 2: Coverage of WM-GM interface**

This metric assesses completeness of the tractography reconstruction in each motor area. The coverage of the WM-GM interface has previously been applied in several studies to quantify tractography performance (Jarbo et al., 2012; Shastin et al., 2021; St-Onge et al., 2018; Wu et al., 2020). The coverage percentage was measured as the number of voxels in the WM-GM interface that were intersected by the streamlines, divided by the total number of voxels in the WM-GM interface. The coverage was measured for the entire motor cortex WM-GM interface and for the WM-GM interface of each motor area. Finally, each coverage measure was reported as the mean value across the left and right hemispheres. **Statistical analysis:** One-way repeated measures analysis of variance (ANOVA) was computed across the tractography methods to determine whether their coverage of the entire motor cortex WM-GM interface was statistically different. Post-hoc paired t-tests were then performed between each pair of the six tractography methods with FDR correction (across 15 comparisons). Then, within each tractography method, one-way repeated measures ANOVA was applied to assess whether WM-GM interface coverage differed across anatomical motor areas. Post-hoc paired t-tests were then performed to compare the coverage between each pair of anatomical motor areas, with FDR correction across the 6 comparisons.

**Metric 3:  Anatomical distribution of CST streamlines (percentage within each anatomical subdivision)**

This metric assesses the percentage of streamlines reconstructed within each subdivision to evaluate the overall anatomy of the CST. The percentage of streamlines was calculated as the number of streamlines from each motor area divided by the total number of CST streamlines in each hemisphere. This metric was reported as the mean value across left and right hemispheres. **Statistical analysis:** Within each tractography method, one-way repeated measures ANOVA was applied to assess whether the percentage of streamlines differed across anatomical motor areas. Post-hoc paired t-tests were then performed to compare the streamline percentage between each pair of anatomical motor areas, with FDR correction across the 6 comparisons. Then, we assessed the relationship between the reconstructed streamlines and the volume of the corresponding motor area. To this end, for each motor area we computed Pearson's correlation between the percentage of streamlines and the percentage of total cortical volume of the corresponding motor area (where both correlated measures were reported as the mean across hemispheres). The correlation coefficient and p-value were reported for each method.

**Considerations for fair comparison across methods**

Tractography methods are known to be sensitive to the seeding strategy used to initiate tractography streamlines (Côté et al., 2013; Girard et al., 2014; K. Schilling et al., 2018). To enable best performance of each tractography method for fair comparison, we performed an experiment to compare four commonly used seeding strategies (including: seeding from the whole brain, white matter, gray matter, and WM-GM interface, as described Supplementary Material 1) for each toolkit (except the tractography method from TractSeg software, which used its own seeds predicted by a deep learning model for each subject). The best-performing strategy for each method, according to the WM-GM interface coverage metric, was then used for all further experiments. For each method, the best-performing strategy was: whole brain seeding strategy for the UKF tractography framework, white matter seeding strategy for the two deterministic tractography methods (SD-Stream and GQI-Det), and WM-GM interface seeding strategy for the iFOD2 method.

To perform tractography, the tractography parameter combination used by each tractography method was determined according to published studies. When selecting parameters, we gave a high priority to parameter combinations recommended in studies published by the toolkit developers. The final selected parameter combinations for each tractography method are provided in Supplementary Table 1. Furthermore, the same total number of streamlines was used for evaluation of all methods. Across all methods, a total of 5,000 streamlines per hemisphere was determined to provide stable results (Supplementary Material 2). If a method resulted in over 5,000 streamlines after selection, random sampling without replacement was used to obtain 5,000 streamlines.

To provide an unbiased comparison across methods, we performed an experiment to determine the most appropriate dataset (single- data or multi-shell data) for each tractography method. These two datasets each have specific properties for streamline reconstruction. The multi-shell data can distinguish the cerebrospinal fluid (CSF), gray matter (GM) and white matter (WM) according to their response functions under different b-values, and several high-angular-resolution diffusion imaging (HARDI) methods have demonstrated that fiber orientations can be estimated with better accuracy from multi-shell data (Assemlal et al., 2011; Cheng et al., 2010; Descoteaux et al., 2011; Jeurissen et al., 2014). The single-shell data with a high b-value (b-value of 3000 s/mm$^2$) has good angular resolution and high sensitivity to crossing fibers (Descoteaux et al., 2007; Ning et al., 2015). Complete details for this experiment are provided in Supplementary Material 3. To briefly summarize, two evaluation metrics, the reconstruction rate and the coverage of WM-GM interface, were used to quantify tractography method performance in single- and multi-shell data. Except for the TractSeg method, all methods had higher reconstruction rates and WM-GM coverage using the single-shell b=3000 s/mm$^2$ data. Thus, for the following experimental evaluation we used multi-shell data as input to the TractSeg method, and single-shell b=3000 data as input to all other tractography methods.

## 3. Results

### 3.1 Visual comparison of the CST reconstructed using the six tractography methods

Figure 5(a) provides a visualization of CST reconstruction results across all tractography methods. Three example subjects, with high, low, and typical tractography performance, are selected for visualization. Figure 5(b) provides streamline heatmaps generated from all 100 HCP subjects, overlaid on a T1w template image (MNI152)[6] (Grabner et al., 2006). The value of a voxel in a heatmap represents the number of subjects that have streamlines passing through the voxel. It can be observed that the pathways obtained with UKF2T, iFOD2, and TractSeg tractography methods densely cover most of the CST white matter and the motor cortex, unlike the CST results generated by UKF1T, SD-Stream, and GQI-Det.

---

[6] http://nist.mni.mcgill.ca/?p=858

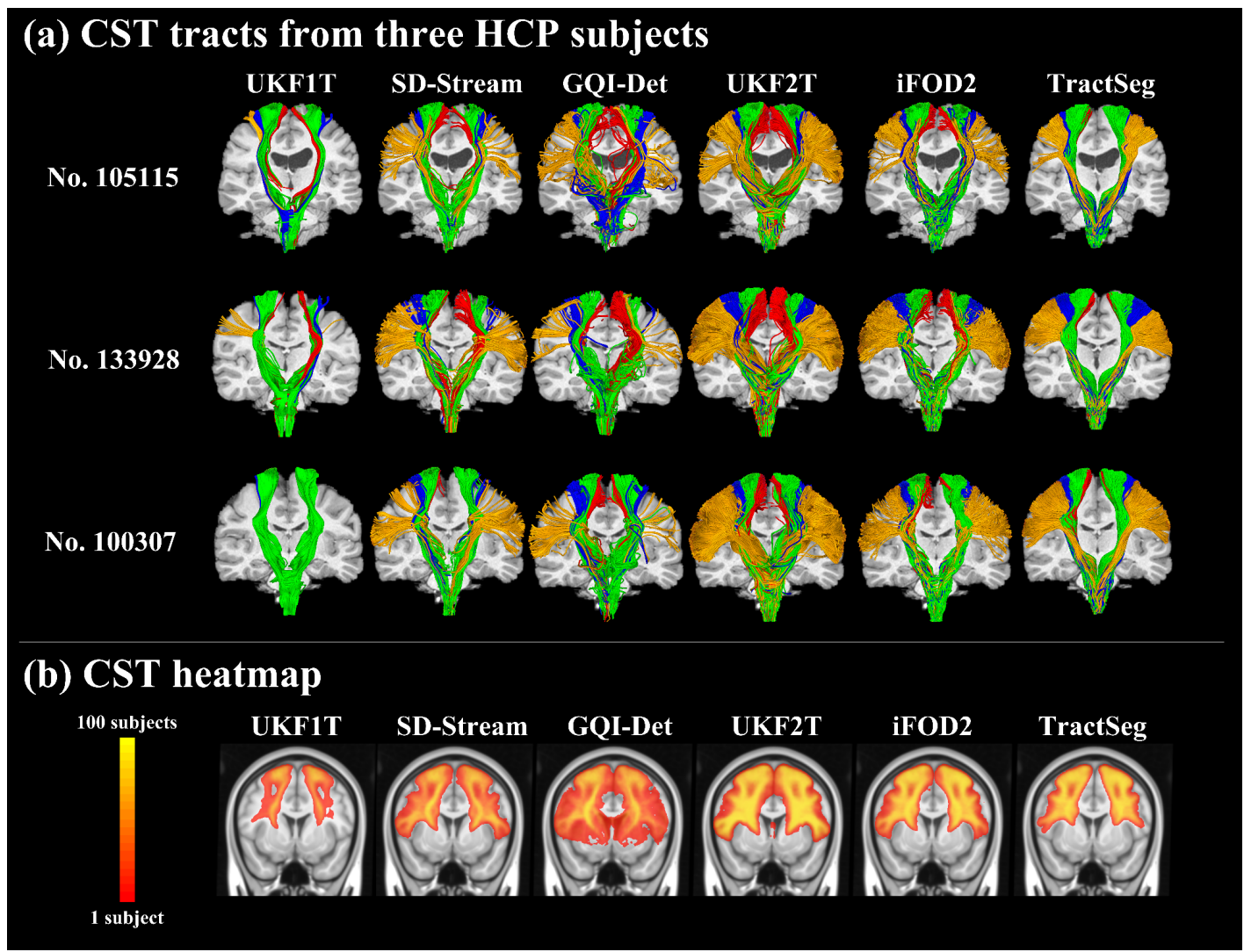

Figure 5. (a) Visual comparison of the CST reconstructed using the six tractography methods. Three subjects are selected as examples: one with low tractography performance (105115), one with high tractography performance (133928), and one with typical tractography performance (100307). Each anatomical subdivision is visualized in a specific color (face: orange, hand: blue, trunk: green, leg: red). (b) Voxel-based tract heatmaps of CST streamlines based on the 100 HCP subjects. The value of a voxel in the heatmaps represents the number of subjects that have fibers passing through the voxel.

### 3.2 CST reconstruction rate

Figure 6 provides the reconstruction rate of the complete CST pathway, i.e. the percentage of subjects where all eight anatomical subdivisions were successfully reconstructed. The reconstruction rate was significantly different across the six tractography methods. The CST reconstruction rate of UKF2T (100%) was significantly higher than the other five tractography methods. This was followed by iFOD2, with a CST reconstruction rate of 79%. As expected, all multi-fiber models outperformed the single-fiber model method (UKF1T), which was unable to reconstruct the complete CST.

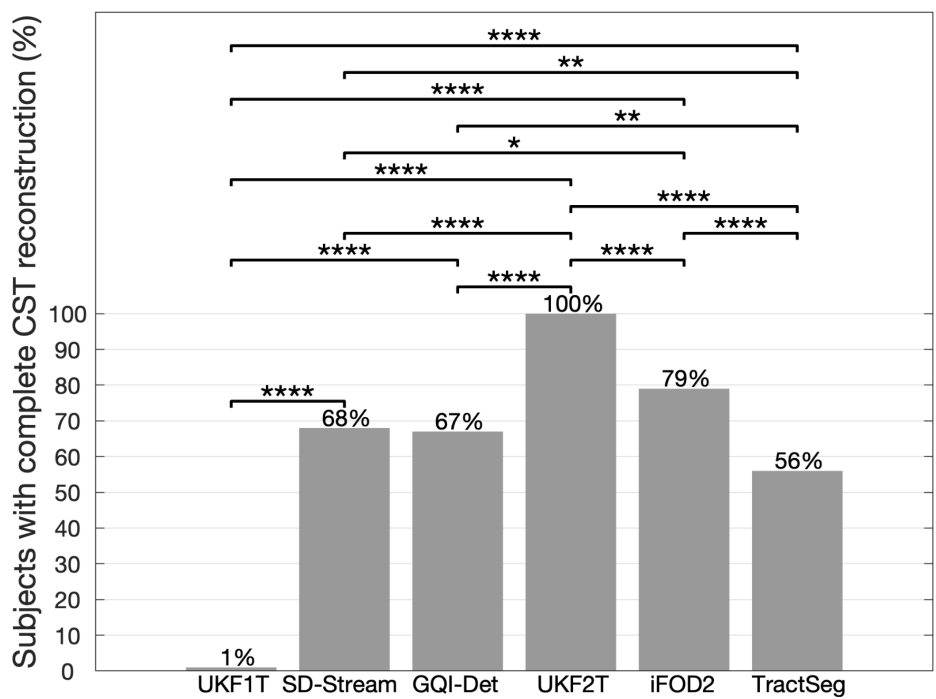

Figure 6. The reconstruction rate of the complete CST was significantly different across the six compared tractography methods (ANOVA, p<0.0001). Post-hoc two-group McNemar's tests with significant results are indicated by asterisks. *: p<0.05; **: p<0.01; ****: p<0.0001. Each bar indicates the mean value.

Figure 7 shows the reconstruction rate of each anatomical subdivision. All tractography methods could successfully reconstruct streamlines (reconstruction rate 100%) in the trunk motor area, while the streamlines in the leg motor area were the most challenging to reconstruct for most of the tractography methods. The UKF1T, SD-Stream, iFOD2, and TractSeg tractography methods produced the lowest reconstruction rates (26%, 72%, 79%, and 50%, respectively) in the left leg motor area, while GQI-Det produced the lowest reconstruction rate (67%) in the right leg motor area. With the exception of UKF1T, all methods were successful in reconstructing the hand and face subdivisions. The SD-Stream, UKF2T, iFOD2, and TractSeg methods produced 100% reconstruction of the hand and face subdivisions, while the GQI-Det method produced reconstruction rates of 94%, 100%, 95%, and 98% for the left face, right face, left hand, and right hand, respectively.

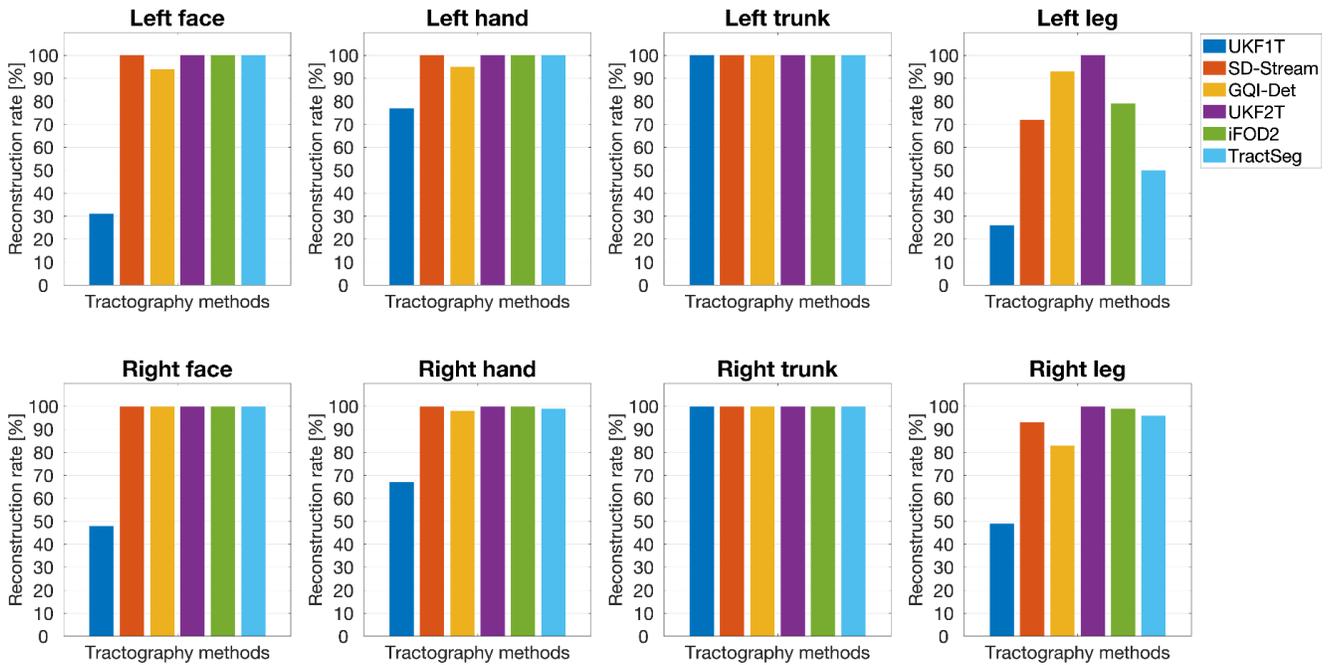

Figure 7. Reconstruction rate of each anatomical subdivision using different tractography methods. Each bar indicates the mean value.

### 3.3 Coverage of WM-GM interface

Figure 8 shows the WM-GM interface coverage of the CST (i.e. the percentage of voxels in the WM-GM interface that were intersected by the CST streamlines). Overall, all methods achieved partial coverage of the WM-GM interface. The UKF2T method obtained the highest coverage of 43.98%, followed by iFOD2 with coverage of 37.05% and TractSeg with coverage of 22.43%. The CST streamlines obtained by the UKF1T, SD-Stream, and GQI-Det methods had low coverage (under 20%). The WM-GM interface coverage results were significantly different across tractography methods (p<0.0001).

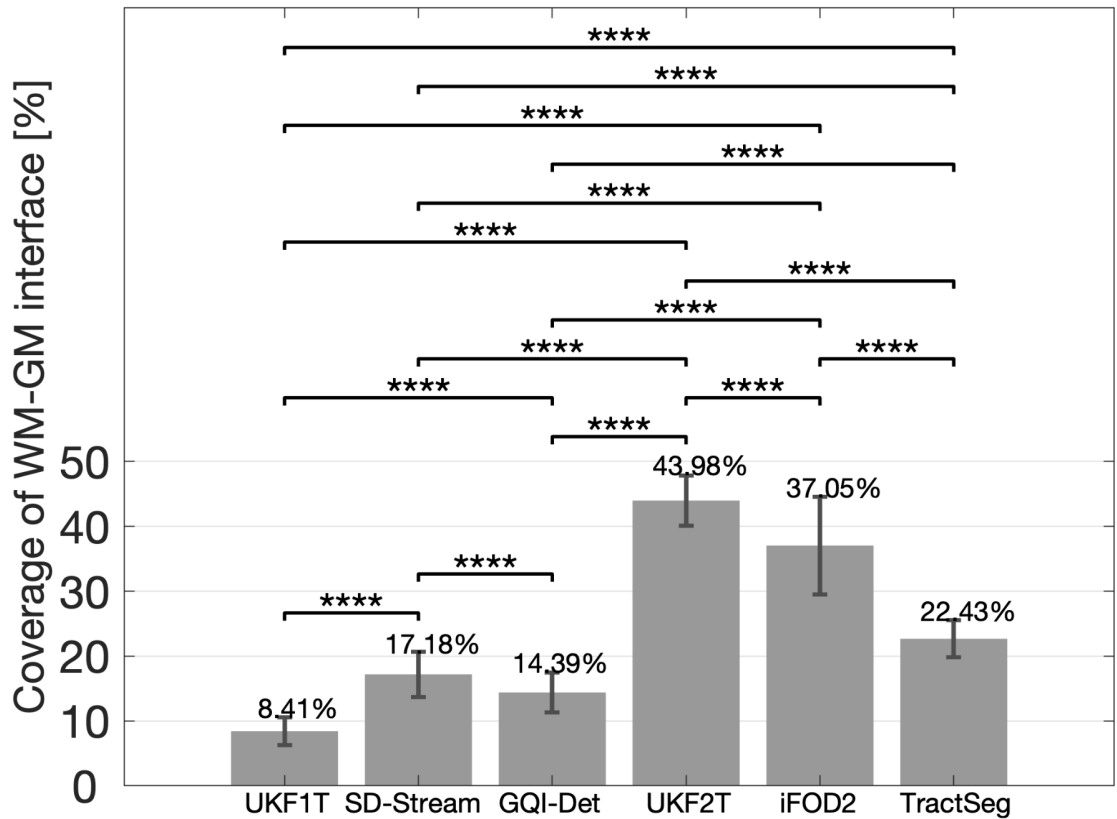

Figure 8. WM-GM interface coverage of the CST. The coverage of the WM-GM interface was significantly different across the six compared tractography methods (ANOVA, p<0.0001). Post-hoc two-group t-tests with significant results are indicated by asterisks. ****: p<0.0001. Each bar indicates the mean value, and the error bar indicates the corresponding standard error.

Figure 9 shows the WM-GM interface coverage of each anatomical subdivision. Overall, for all methods, there were significant differences in WM-GM coverage across the different anatomical subdivisions of the CST. The highest WM-GM interface coverage (ranging from 35.10%-71.66% across methods) was achieved in the trunk motor area. For all methods, the WM-GM interface coverage of the trunk motor area was significantly higher than the coverage of the face, leg, and hand areas, which were generally low. The highest coverage of the face area (39.79%) was achieved by the UKF2T method, followed by the iFOD2 method (26.72%), while the highest coverage of the hand area (48.05%) was achieved by the iFOD2 method, followed by the UKF2T method (37.81%). Finally, the highest coverage of the leg area (31.64%) was achieved by the UKF2T method, followed by the iFOD2 method (15.59%).

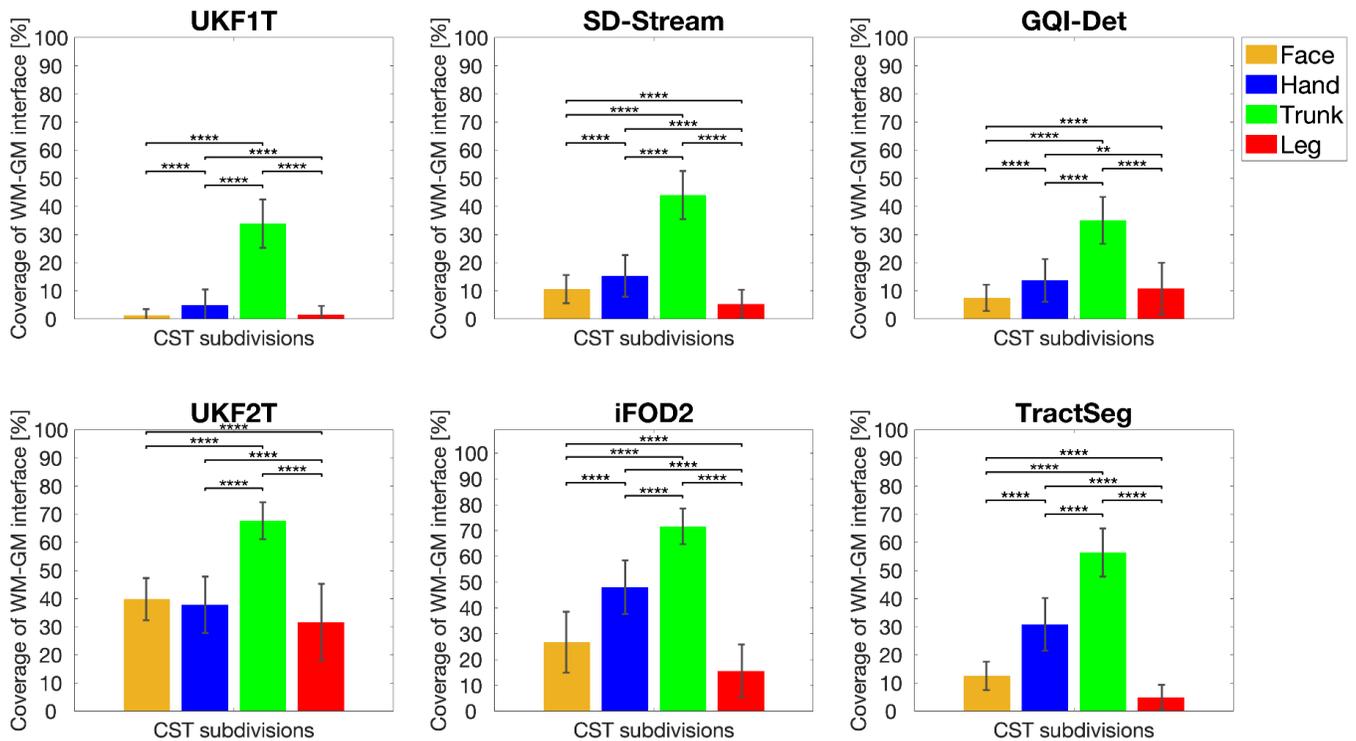

Figure 9. WM-GM interface coverage of the anatomical subdivisions. For all tractography methods, the coverage of the WM-GM interface was significantly different across the four anatomical subdivisions (ANOVA, $p<0.0001$). Post-hoc two-group t-tests with significant results are indicated by asterisks. **: $p<0.01$, ****: $p<0.0001$. Each bar indicates the mean value, and the error bar indicates the corresponding standard error.

Figure 10 depicts a cortical surface map of CST termination points for all tractography methods. One example subject is selected for visualization. It can be observed that the UKF2T, iFOD2, and TractSeg methods have higher WM-GM interface coverage than UKF1T, SD-Stream, and GQI-Det. (This also can be observed in a coronal view in Supplementary Figure 6). However, it can be seen that all of the tractography methods generate CST streamlines that prefer to terminate on gyral crowns, rather than the banks of sulci. In fact, even with exhaustive streamline seeding (approximately 30 million seeds, or over 1,600 seeds per WM-GM interface voxel), it is still not possible to obtain complete cortical coverage (as shown in Supplementary Figure 7).

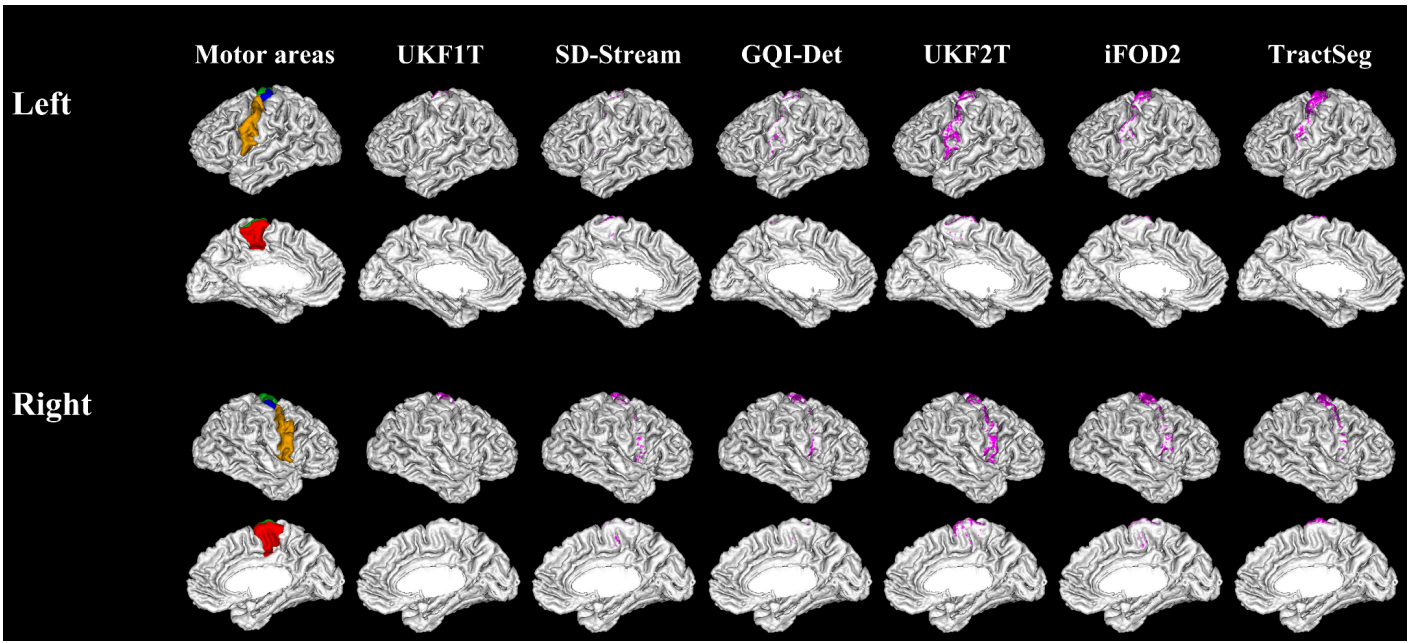

Figure 10. Visualization of the white matter-gray matter interface surface including streamline termination points for each tractography method. Pink regions indicate at least one fiber termination point, while white regions indicate no fibers passed. Note the pink regions are mainly seen on gyral crowns, while sulcal regions are predominantly shown in white. One typical subject (100307) is selected. Each anatomical subdivision is visualized in a specific color (face: orange, hand: blue, trunk: green, leg: red).

### 3.4 Anatomical distribution of streamlines and correlation with cortical volumes

Figure 11(a) shows the percentage of motor cortex volume within each motor cortical area. It is clear that the face motor cortical area occupies the largest percentage of the motor cortex (53.79%). Figure 11(b) shows the percentage of streamlines reconstructed within each anatomical subdivision, i.e. the overall anatomical distribution of the CST streamlines, for each tractography method. While an ideal reconstruction would be expected to have a high percentage of streamlines in the face motor area (because this area has the highest cortical volumes), the highest percentage of streamlines (47.52%-87.34%) was obtained within the trunk motor area for all tractography methods. In comparison, the percentage of streamlines within the face motor area ranged from 1.79% to 34.21%. Overall, for all methods, the percentage of streamlines within the trunk motor area was significantly higher than all other anatomical subdivisions.

Figure 12 provides the correlation between the anatomical distribution of streamlines and the cortical volumes of each motor area. An ideal reconstruction would be expected to have a strong positive correlation between these two measures. However, due in large part to the bias toward reconstructing trunk streamlines, almost all methods produced a negative correlation between the percentage of streamlines and the percentage of cortical volume corresponding to each anatomical subdivision. This phenomenon was mitigated in the UKF2T method, which was the only method to produce a positive correlation (r=0.3757, p<0.0001).

.

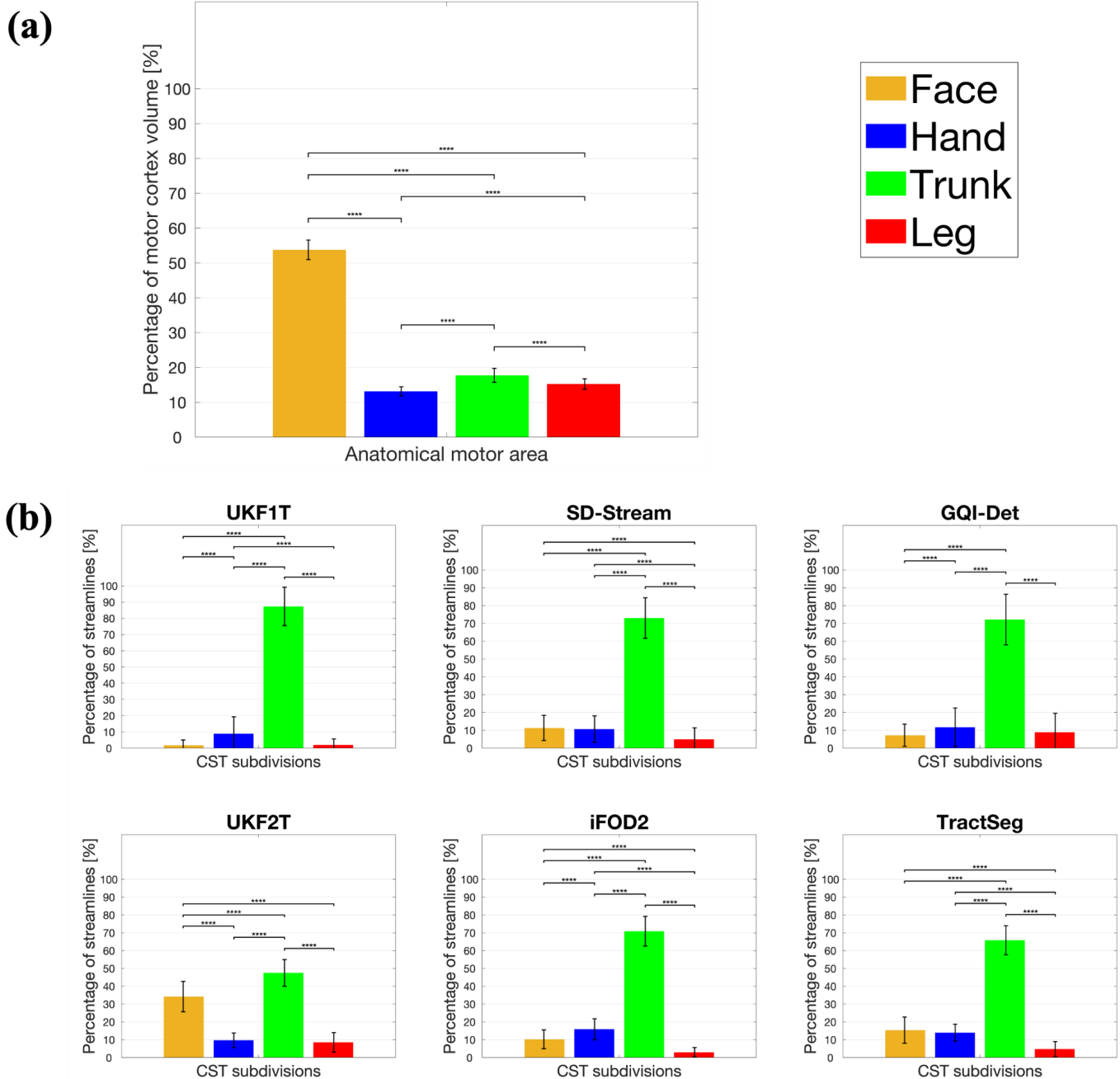

Figure 11. (a) The percentage of motor cortical volume in each motor area. All percentages were significantly different across the four motor areas (ANOVA, $p<0.0001$). Post-hoc pairwise t-tests with significant results are indicated by asterisks. ****: $p<0.0001$. Each bar indicates the mean value, and the error bar indicates the corresponding standard error. (b) Anatomical distribution of streamlines across anatomical subdivisions for each tractography method. For all tractography methods, the percentage of streamlines was significantly different across the four anatomical motor areas (ANOVA, $p<0.0001$). Post-hoc pairwise t-tests with significant results are indicated by asterisks. ****: $p<0.0001$. Each bar indicates the mean value, and the error bar indicates the corresponding standard error.

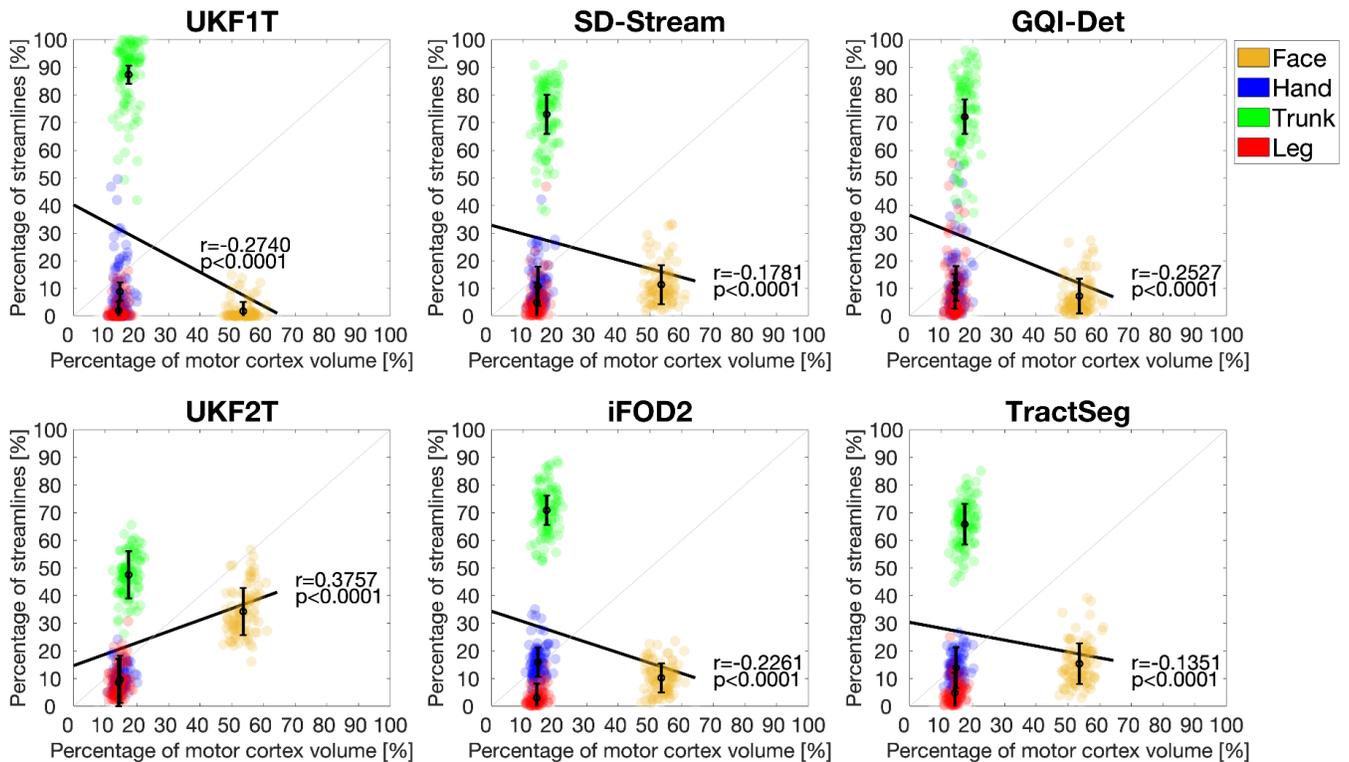

Figure 12. Correlations between the percentage of streamlines and the percentage of volume in each motor area. The correlation coefficient of each tractography method is reported.

## 4. Discussion

In this study, we explored the performance of several tractography methods for reconstructing the CST and its somatotopic organization. We compared six tractography methods for CST reconstruction: GQI-based deterministic tractography (GQI-Det), CSD-based deterministic (SD-Stream) and probabilistic (iFOD2) tractography, UKF tractography with one-tensor (UKF1T) and two-tensor (UKF2T) models, and TractSeg. We used four quantitative measurements including reconstruction rate, the WM-GM interface coverage, anatomical distribution of streamlines and correlation with cortical volumes to assess the advantages and limitations of each method. We made several overall observations, as follows. First, results showed that no tractography method produced a good correlation between quantitative measures of connecting streamlines and volumes of related cortical motor areas. Second, all of the tractography methods had a bias toward reconstructing trunk streamlines (ranging from 35.10%-71.66% of total streamlines across methods). Third, all methods generated low coverage of the WM-GM interface in the largest motor area (face). These observations demonstrate that while modern tractography methods have made progress towards the well-known challenge of improving the reconstruction of the lateral projections of the CST (Pujol et al., 2015), the overall problem of performing an comprehensive CST reconstruction, including both lateral (hand and face) and medial portions (i.e. leg motor area), remains an important challenge for dMRI tractography. Below we discuss some detailed observations about our results in comparison with related studies in the literature.

As expected, we demonstrated that multi-fiber tractography (SD-Stream, iFOD2, UKF2T, GQI-Det, and TractSeg) has advantages for reconstructing the complete CST compared to a single-fiber model (UKF1T). This is likely because single-fiber models cannot estimate multiple fiber orientations and thus cannot effectively track in the region of multiple crossing fibers (e.g. the centrum semiovale), which is a well-known issue in the tractography literature (Behrens et al., 2007; Jeurissen et al., 2019; Zhang et al., 2022). Thus, the UKF1T method obtained the lowest reconstruction rate in the leg, hand and face motor areas. Multiple studies have previously shown that multi-fiber models are beneficial for CST reconstruction (Behan et al., 2017; D. Q. Chen et al., 2016; Z. Chen et al., 2016; Xie et al., 2020).

We found that seeding strategies for tractography can highly affect CST reconstruction performance. Previous studies have shown that different seeding strategies can have a significant impact on the performance of tractography algorithms (Côté et al., 2013; Girard et al., 2014; St-Onge et al., 2021; Theberge et al., 2022). By testing multiple seeding strategies, in this study we identified an optimal seeding strategy for CST reconstruction for each tractography method. We determined that the optimal seeding strategy depends on the specific algorithm being used. In fact, the two methods with the highest white matter-gray matter interface coverage (UKF2T and iFOD2) required different seeding strategies for best performance (whole-brain and WM-GM interface seeding, respectively). This is likely due to the different ways in which these algorithms propagate streamlines.

We found that no tractography method could achieve a good correlation between the percentage of connecting streamlines and the percentage of total cortical volume of the corresponding motor area. Neuroanatomical research across multiple primate species shows that the lateral motor areas (corresponding to hand and face) actually have higher neuron density than the more medial motor areas (Young et al., 2013). This implies that an ideal CST reconstruction would obtain even more streamlines to the hand and face areas than would be expected from the volume of these areas alone. However, instead of producing large numbers of streamlines to the hand and face areas, all tractography methods were shown to have a bias toward reconstructing trunk streamlines. This bias is driven in part by challenges in modeling the crossing of multiple fibers, challenges in tractography when multiple fibers share the same parallel orientation (bottleneck issue), and challenges in reconstructing curved fibers (as shown in Figure 2). The most medial CST streamlines connecting to the trunk region are less affected by this multiple fiber modeling challenge. It is also apparent in Figure 2 that streamlines must curve to connect to the lateral motor areas. However, line propagation methods have bias toward reconstructing simple trajectories and against reconstructing high curvature tracts (Jeurissen et al., 2019; Roebroeck et al., 2008), and fiber tracking may stop in regions where streamline curvature exceeds a threshold.

We found that no tractography method under study could generate good WM-GM interface coverage, as all tractography algorithms obtained overall coverage of under 50%. As visualized in Figure 10, this is in part due to the termination of CST streamlines at brain gyri rather than sulci (the gyral bias problem). From the perspective of tracking strategy, gyral bias relates to the preference of tractography algorithms to follow the path with least angular deviation, rather than making the sharp turns necessary to exit the white matter (K. Schilling et al., 2018). Another reason for the low WM-GM interface coverage is that there are fewer streamlines that terminate in the lateral motor areas. While WM-GM interface coverage can be improved by simply increasing the number of seeded streamlines, there is a plateau in the WM-GM interface coverage that can be achieved, and 100% coverage cannot be reached (Supplementary Figure 7) (St-Onge et al., 2021). Over-representation of easy-to-track parts of bundles is still a critical problem for current tractography methods.

Furthermore, we note that all methods except TractSeg achieved significantly higher coverage using single-shell b=3000 data, in comparison with multi-shell data (Supplementary Material 3). This indicates that future work on tractography method design for CST reconstruction should consider fully leveraging the rich information provided in multi-shell dMRI data.

There is a trade-off between sensitivity and specificity in tractography when reconstructing the CST. The UKF2T, iFOD2 and TractSeg methods had higher coverage performance than the other three tractography methods. The CST reconstructed by UKF2T obtained the highest WM-GM interface coverage (about 44%) indicating high sensitivity, while the TractSeg method obtained about 22% coverage, indicating relatively lower sensitivity. But as can be observed in Figure 5(a), the apparent higher sensitivity of UKF2T results in more visually apparent false positive streamlines in comparison with the TractSeg results. TractSeg constrains the tracking region to ensure relatively few false-positive streamlines, in comparison with other methods shown in Figure 5(a). Finding a balance between sensitivity and specificity is a well-known challenge in dMRI (Alexander et al., 2019; Novikov et al., 2019) and will be crucial to increase the accuracy and reliability of CST reconstruction results.

Potential limitations of the present study, including suggested future work to address limitations, are as follows. First, in this study, we chose the widely used WMQL method (Wassermann et al., 2016) to select CST streamlines connecting the brainstem and precentral cortex. However, this approach ignores the accuracy of the CST pathway in the white matter. Future work may require the addition of ROIs such as the posterior limb of the internal capsule to reduce false positives for CST reconstruction. Second, in this study we focused on the somatotopic organization of the CST at the level of the motor cortex. Future work can compare performance of tractography methods for depicting somatotopy at the level of the brainstem or the posterior limb of the internal capsule. Finally, we focused on CST tracking using dMRI data from healthy young adults in the current study. Future work could investigate tracking performance in dMRI data across the lifespan (Harms et al., 2018) or in the context of neurosurgical planning (Essayed et al., 2017b). Furthermore, future multimodal neuroimaging studies integrating fMRI with dMRI are needed to add insight on new understandings regarding the functional neuroanatomy of M1, which has been recently indicated to be involved in the integration of motricity with goal-oriented cognitive control (Gordon et al., 2023).

## 5. Conclusion

The main contribution of this work is to comprehensively investigate the performance of multiple tractography methods for reconstructing the CST and its somatotopic organization at the level of the motor cortex. We found that the multi-fiber UKF tractography (UKF2T) method has the best performance, obtaining the highest reconstruction rate and cortical coverage. Overall, we conclude that while current tractography methods have made progress towards the well-known challenge of improving the reconstruction of the lateral projections of the CST, the overall problem of performing an comprehensive CST reconstruction including both lateral (hand and face area) and medial portions (leg area) remains an important challenge for dMRI tractography.

## Acknowledgements

We gratefully acknowledge funding provided by the following grants: National Natural Science Foundation of China grants 61976190; National Institutes of Health (NIH) grants R01MH125860, R01MH119222, P41EB015902, P41EB028741, R01NS125307, R01NS125781, R01MH112748, R01MH111917,

## Data availability statement

Imaging datasets of 100 subjects from the Human Connectome Project (https://www.humanconnectome.org) are used in this paper. The datasets are available online. The computed CST tractography data will be made available on request.

## Conflict of interest disclosure

The authors have no conflict of interest.

# Supplementary Material to: Comparison of multiple tractography methods for reconstructing the somatotopic organization of the corticospinal tract

## 1. Experiment to determine seeding strategy for each tractography method

Tractography methods are known to be sensitive to the seeding strategy used to initiate tractography streamlines (Côté et al., 2013; Girard et al., 2014; K. Schilling et al., 2018). To enable best performance of each tractography method for fair comparison, we performed an experiment to compare four commonly used seeding strategies (as shown in Figure 3) for each toolkit (except the tractography method from TractSeg software, which used its own seeds predicted by a deep learning model for each subject). The four compared seeding strategies are widely used in tractography research (Calamante et al., 2010; Clayden, 2013; Girard et al., 2014; K. Schilling et al., 2018). *Seeding strategy 1* seeded tractography throughout all voxels within the tractography mask (Calamante et al., 2010; Radmanesh et al., 2015). *Seeding strategy 2* seeded tractography throughout the white matter (Côté et al., 2013; Radmanesh et al., 2015). *Seeding strategy 3* seeded tractography in gray matter (Reveley et al., 2015). *Seeding strategy 4* seeded from the interface of the white matter-gray matter (WM-GM) boundary (Girard et al., 2014; Parker et al., 2014; Smith et al., 2012).

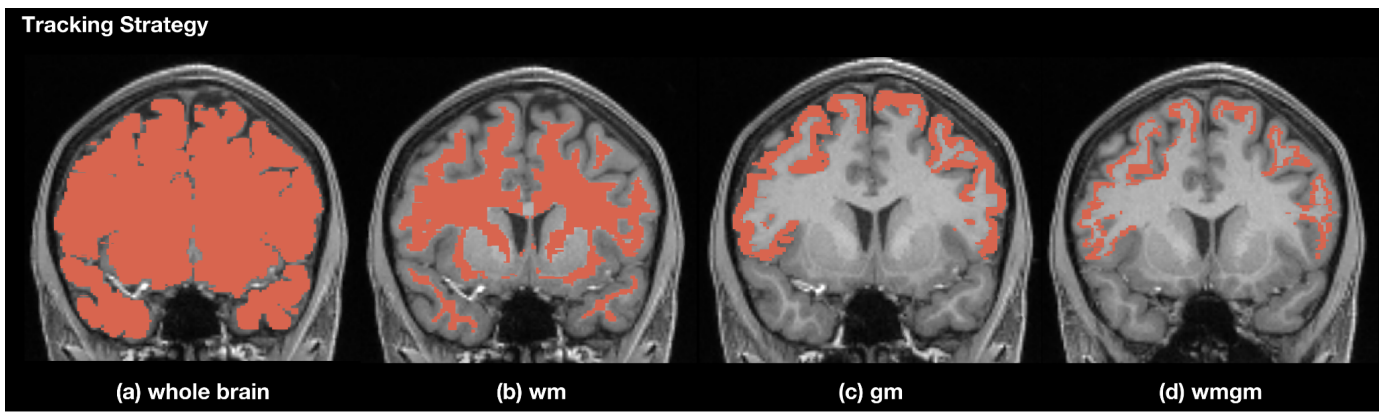

Supplementary Figure 1. Regions employed for tractography seeding strategies. (a) seeding in the whole brain within the mask for tractography. (b) seeding in the white matter. (c) seeding in the gray matter. (d) seeding in the WM-GM interface.

All seeding strategies were tested using data from ten randomly selected HCP subjects. For fair comparison across the four seeding strategies, after ROI selection we retained 2,000 streamlines for each brain hemisphere and for each tractography method. The number of 2,000 streamlines was chosen for reasonable computation time and memory use when performing tractography. CST selection was performed as described in Section 2.3. The performance of each method was quantified using the WM-GM interface coverage metric (the percent of WM-GM interface voxels intersected by at least one streamline, as described in Section 2.4). Further experiments used the best-performing seeding strategy for each method.

Supplementary Figure 2 shows the results of the seeding strategy experiment. These results indicate that different tractography methods have different optimal seeding strategies. The highest coverage of the WM-GM interface (34.80%) was obtained by the UKF2T method using a whole brain seeding strategy, closely followed by the UKF2T coverage result using a white matter seeding strategy (34.63%). The next highest coverage (23.68%) was obtained by the iFOD2 method using a WM-GM interface seeding strategy. The deterministic tractography methods with multi-fiber models (MSMT-CSD+SD-Stream and GQI-Det) produced the highest coverage using a white matter seeding strategy, while the single fiber model method (UKF1T) produced the highest coverage using a whole-brain seeding strategy.

**(a) Data of b=3000 s/mm²**

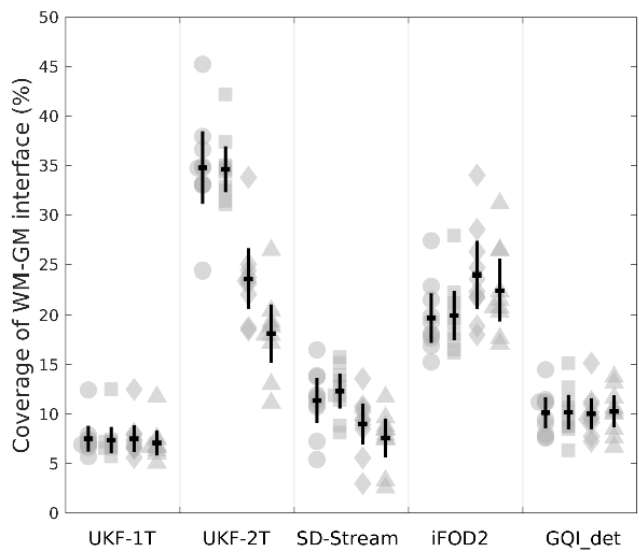

**(b) Multi-shell data**

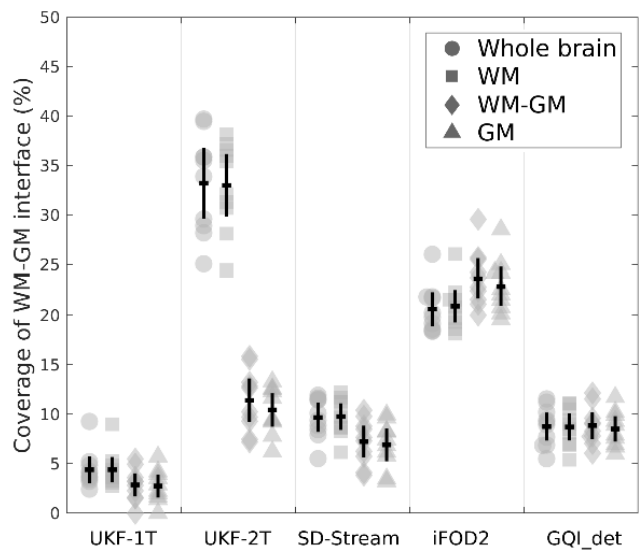

Supplementary Figure 2. The coverage of the WM-GM interface that is intersected by each tractography method, using different seeding strategies. Four seeding strategies are compared: whole brain (blue), white matter (red), WM-GM interface (yellow), gray matter (purple).

## 2. Experiment to determine a total number of streamlines for stable subdivision identification

**Experimental details:** Data (single- and multi-shell) from ten randomly selected HCP subjects was studied. For each tractography method, we used the optimal seeding strategy based on the results from Supplementary Material 1. After exhaustive seeding by each method, the total number of retained streamlines was varied, and the percentage of streamlines connecting to each motor area was computed. We tested 1,000 to 10,000 (increment of 1,000) retained streamlines using random sampling. Supplementary Figures 1 and 2 show the measurement of the percentage.

**Results:** Supplementary Figure 3 gives the results of the percentage of streamlines in each anatomical subdivision using different tractography methods. Across all methods, a total of 5,000 streamlines was determined to provide a very stable measurement of the percentage of streamlines present in each anatomical subdivision (left face, left UE, left trunk, left LE, right face, right UE, right trunk, and right LE). Therefore, in the rest of the manuscript, to fairly and effectively compare the tractography methods, we retained 5,000 streamlines for each brain hemisphere and for each tractography method, with a fiber length threshold of 70mm to eliminate any effect from fibers too short to form part of the CST anatomy. If a method resulted in over 5,000 streamlines after selection, random sampling without replacement was used to obtain 5,000 streamlines.

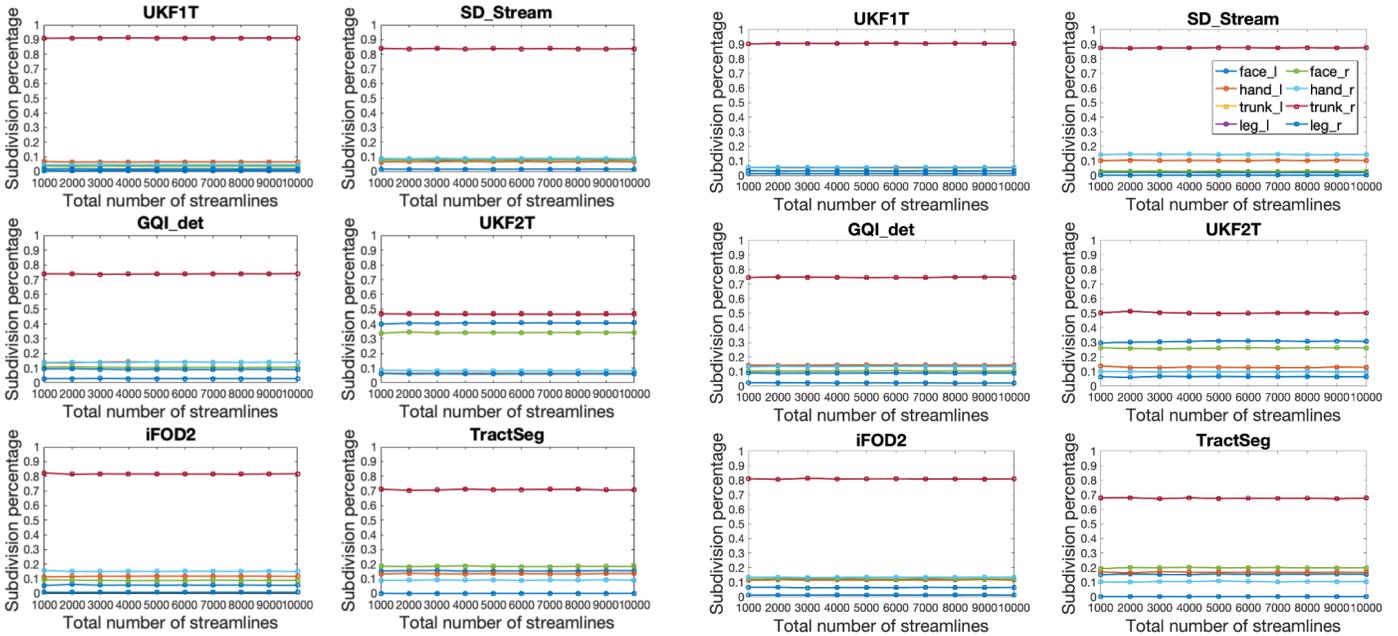

Supplementary Figure 3. Percentage of streamlines in each anatomical subdivision using different tractography methods for (a) data of b=3000 s/mm² and (b) multi-shell data. Total number of streamlines ranges from 1,000 to 10,000.

### 3. Experiment to determine dataset for each tractography method

Data (single- and multi-shell) from 100 selected HCP subjects was studied. For each tractography method, we used the optimal seeding strategy based on the results from Supplementary Material 1 and each hemisphere generated a total number of 5,000 CST fibers for comparison (based on Supplementary Material 2) and tractography parameters for each method are based on Supplementary Table 1. Then, we computed the reconstruction rate of the complete CST and the coverage of the WM-GM interface for each CST reconstruction result. A statistical comparison was then performed across the single- and multi-shell data. For the reconstruction rate of complete CST results, McNemar's tests (McNEMAR, 1947) were performed (a total of 6 comparisons). For the coverage of WM-GM interface results, paired t-tests were performed.

Supplementary Figure 4 gives the results of the reconstruction rate of complete CST, i.e. the percentage of subjects where all eight anatomical subdivisions were successfully reconstructed. For SD-Stream and GQI-Det, the reconstruction rate results from the single-shell data are significantly higher than results from the multi-shell data. The iFOD2 method has better reconstruction rate on single-shell data than multi-shell data as well but without significance. The TractSeg method has a significantly better reconstruction rate when performed on the multi-shell data. Supplementary Figure 5 gives the results of the coverage of WM-GM interface. Except for TractSeg, all other tractography methods obtain significantly higher coverage on the single-shell data than multi-shell data. Overall, TractSeg has better CST performance on multi-shell data, and all other tractography methods have better CST performance on single-shell data.

| Supplementary Table 1. Tractography parameters for each method | | |
|---|---|---|
| **Methods (model + tractography)** | **Parameters** | **Supporting studies** |
| CSD (or MSMT-CSD) + iFOD2 | StoppingCutoff = 0.1, Angle = 45 (default settings for MRtrix 3.0.2) | (Bauer et al., 2020; Schiavi et al., 2020; Zhylka et al., 2020) |
| CSD (or MSMT-CSD) + SD_Stream | StoppingCutoff = 0.1, Angle = 60 (default settings for MRtrix 3.0.2) | (Barakovic et al., 2021; Schiavi et al., 2020) |
| GQI + Det | default anisotropy threshold (0.6 * Otsu's threshold) for terminating streamlines, Max angle = 70 | (Verstynen et al., 2011; F.-C. Yeh et al., 2018) |
| UKF1T | seedingFA = 0.1 stoppingFA = 0.08, stoppingThreshold = 0.06 | (Zhang, Wu, et al., 2018) |
| UKF2T | seedingFA = 0.1 stoppingFA = 0.08, stoppingThreshold = 0.06 | (Zhang, Wu, et al., 2018) |
| TractSeg | Not applicable | (Wasserthal et al., 2019) |

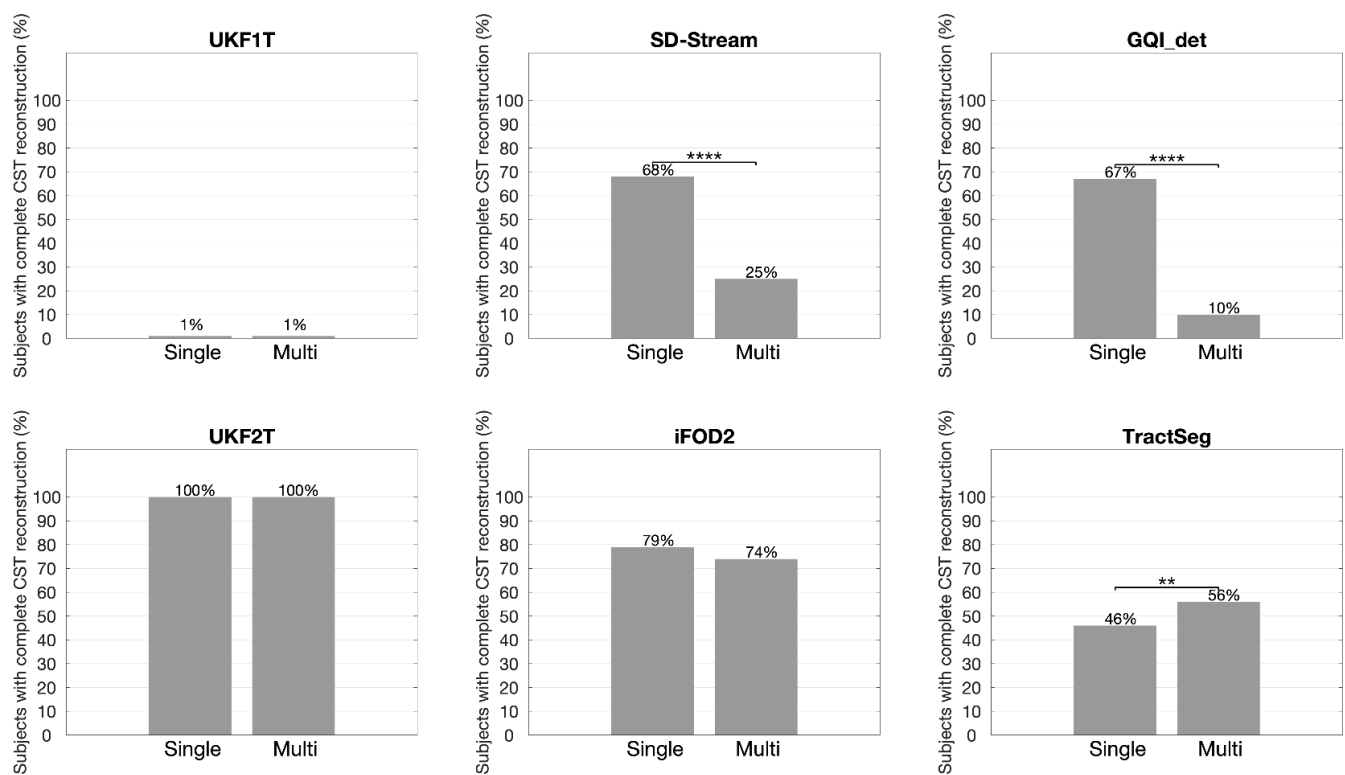

Supplementary Figure 4 Reconstruction rate of the complete CST between two datasets (single- and multi-shell data) for each tractography method. Two-group McNemar's tests with significant results are indicated by asterisks. **: p<0.01; ****: p<0.0001.

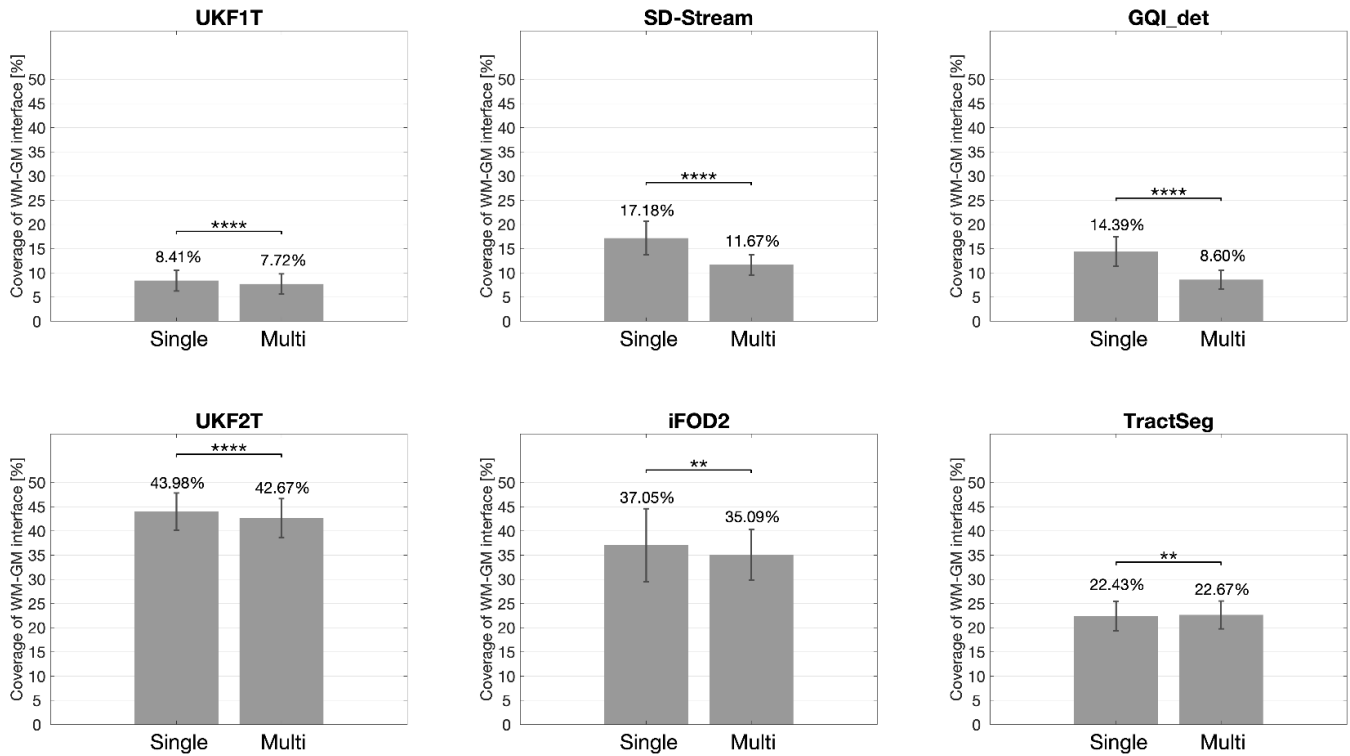

Supplementary Figure 5. The coverage of WM-GM interface between two datasets (single- and multi-shell data) for each tractography method. Paired t-tests with significant results are indicated by asterisks. **: p<0.01; ****: p<0.0001.

## 4. Reconstruction rate of each anatomical subdivision and statistical results between tractography methods

| Supplementary Table 2. Reconstruction rate statistical comparison results for each anatomical subdivision. Except for the anatomical subdivisions of the left and right trunk, the reconstruction rates were statistically significantly different across the six compared tractography methods (Cochran's Q test, p<0.0001) Post-hoc two-group McNemar's tests with significant results are indicated by asterisks. *: p<0.05, **: p<0.01, ***: p<0.001, ****: p<0.0001. | | | | | | | | |
|---|---|---|---|---|---|---|---|---|
| **Compared algorithm pair** | **face-l** | **hand-l** | **trunk-l** | **leg-l** | **face-r** | **hand-r** | **trunk-r** | **leg-r** |
| Overall Cochran's Q result | **** | **** | n.s | **** | **** | **** | n.s | **** |
| UKF1T - UKF2T | **** | **** | - | **** | **** | **** | - | **** |
| UKF1T - SD-Stream | **** | **** | - | **** | **** | **** | - | **** |
| UKF1T - iFOD2 | **** | **** | - | **** | **** | **** | - | **** |
| UKF1T - GQI-det | **** | **** | - | **** | **** | **** | - | **** |
| UKF1T - TractSeg | **** | **** | - | *** | **** | **** | - | **** |
| UKF2T - SD-Stream | - | - | - | **** | - | - | - | * |
| UKF2T - iFOD2 | - | - | - | **** | - | - | - | 1.0000 |
| UKF2T - GQI-det | * | 0.0625 | - | * | - | 0.5000 | - | **** |
| UKF2T - TractSeg | - | - | - | **** | - | 1.0000 | - | 0.1250 |
| SD-Stream - iFOD2 | - | - | - | 0.2100 | - | - | - | * |
| SD-Stream - GQI-det | * | 0.0625 | - | **** | - | 0.5000 | - | * |
| SD-Stream - TractSeg | - | - | - | *** | - | 1.0000 | - | 0.5488 |
| iFOD2 - GQI-det | * | 0.0625 | - | ** | - | 0.5000 | - | **** |
| iFOD2 - TractSeg | - | - | - | **** | - | 1.0000 | - | 0.3750 |
| GQI-det - TractSeg | **** | 0.0625 | - | **** | - | 1.0000 | - | ** |

Abbreviation: *face-l*, *hand-l*, *trunk-l*, and *leg-l* means left face, hand, trunk, and leg; *face-r*, *hand-r*, *trunk-r*, and *leg-r* means right face, hand, trunk, and leg; *n.s.* means no statistical test performed; "-" means both groups' values are exactly the same.

## 5. Visualization of the WM-GM interface coverage across all tractography methods

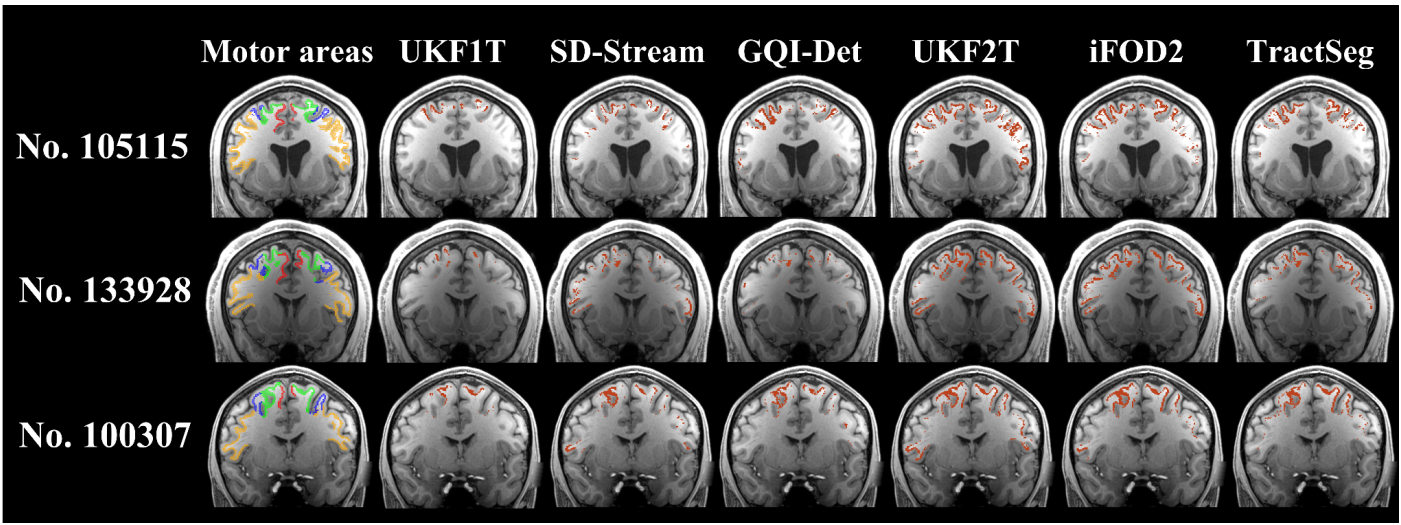

Supplementary Figure 6. Visual comparison of the WM-GM interface coverage using the six tractography methods. Three subjects are selected as examples: one with low tractography performance (105115), one with high tractography performance (133928), and one with typical tractography performance (100307). Each motor area is visualized in a specific color (face: orange, hand: blue, trunk: green, leg: red). For each tractography method column, voxels with color red represent there is at least one fiber passed.

## 6. The relationship between number of streamlines and coverage of WM-GM interface

Increasing the number of streamlines (NoS) will improve the WM-GM interface coverage result. Therefore, we performed an experiment to investigate if obtaining complete WM-GM interface coverage is possible if we generate enough streamlines. To do this, we selected the iFOD2 tractography method, which has low computational cost and can obtain relatively high WM-GM interface coverage (based on Figure 7). We used input dMRI data from 5 subjects. For each subject, we performed exhaustive seeding of the WM-GM interface (a total number of 500,000 CST streamlines was generated, which required using around 30 million seeds per subject). We tested 5,000 to 500,000 (increment of 5,000) retained streamlines using random sampling.

Supplementary Figure 7 gives the relationship between the number of streamlines and WM-GM interface coverage (mean across the 5 subjects tested). As expected, as the number of fibers increases, the measured WM-GM interface coverage increases. However, the relationship is approximately a logarithmic curve, with a leveling off or plateau in the coverage that can be achieved even with increasing numbers of streamlines. Furthermore, even with seeding approximately 30 million seeds, the iFOD2 method can only obtain a WM-GM interface coverage of 73.87%.

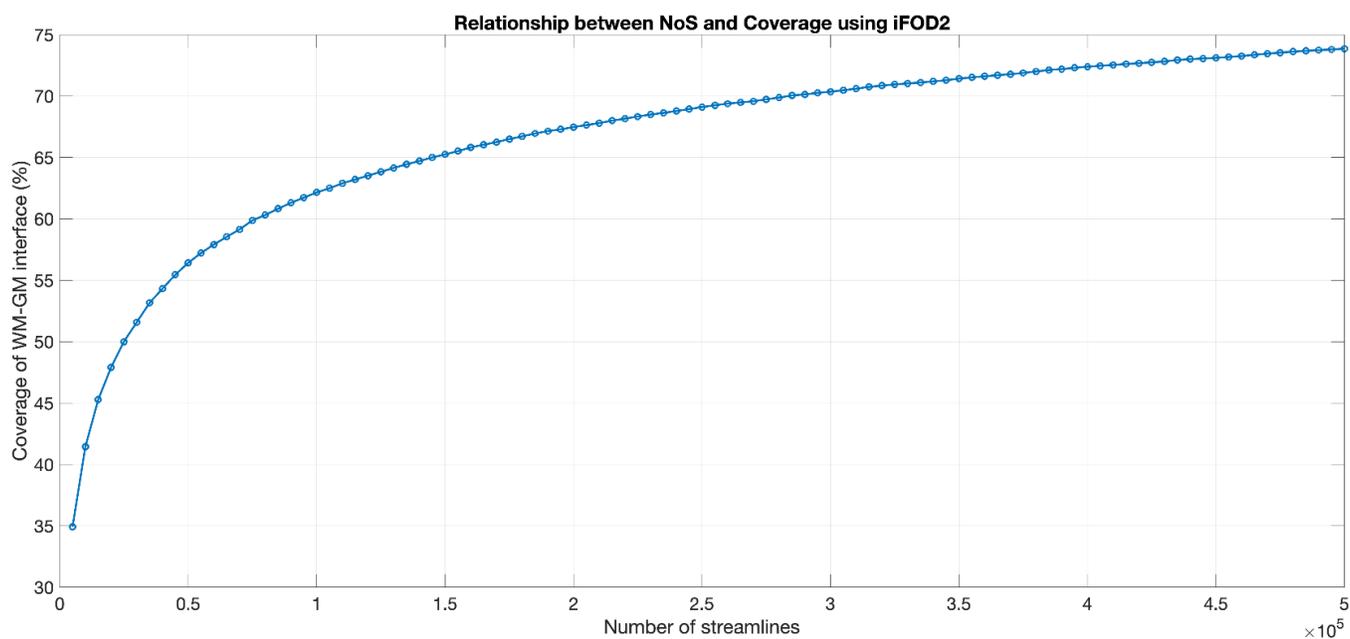

Supplementary Figure 7. The relationship between number of streamlines and coverage of the WM-GM interface using the iFOD2 tractography method.